\documentclass[letterpaper, 10 pt, conference]{ieeeconf}  

\IEEEoverridecommandlockouts                              

\usepackage{amsmath} 
\usepackage{amssymb}  

\usepackage{graphicx}
\usepackage[adjust]{cite}
\usepackage[hidelinks]{hyperref}
\usepackage{url}
\usepackage{xspace}
\usepackage{siunitx}
\usepackage{tikz}
\usepackage{xcolor, colortbl}
\definecolor{Gray}{gray}{0.9}
\usepackage{booktabs}
\usepackage{lipsum}

\title{\LARGE \bf
Designing robot swarms: \emph{a puzzle}, \emph{a problem}, and \emph{a mess} 
}

\author{David Garzón Ramos and Sabine Hauert
\thanks{The authors are with the Bristol Robotics Laboratory (BRL), University of Bristol, United Kingdom. The paper was drafted by DGR and refined by DGR and SH. Mail to 
{\tt\small david.garzonramos@bristol.ac.uk}.} %
\thanks{The research has received funding from the European Innovation Council~(EIC) under the European Union’s Horizon Europe research and innovation programme (EMERGE: grant~No.~101070918). UK participants in the EMERGE project are supported by UKRI (grant~No.~10038942). SH~acknowledges support from the ESRC Centre for Sociodigital Futures (grant~No.~ES/W002639/1).} %
}

\begin{document}
\bstctlcite{IEEEexample:BSTcontrol}

\maketitle
\thispagestyle{empty}
\pagestyle{empty}


\section{INTRODUCTION}\label{sec:intro}
Swarm robotics is the study of how to design self-organized, autonomous groups of robots~\cite{DorBirBra2014SCHOLAR,Ham2018book}.
In a robot swarm, coordinated collective behavior emerges from interactions among robots and between robots and their environment~\cite{Ben2005sab,Sah2005sab}.
Consequently, \emph{designing a robot swarm} has traditionally been associated with identifying or engineering interaction rules to achieve a specific desired collective behavior.
This association has been applied throughout the field~\cite{BraFerBirDor2013SI}, from studying self-organization in laboratory settings~\cite{RubCorNag2014SCI} to devising robot swarms that can help tackle real-world environmental challenges~\cite{TzoSalMcc-etal2024arso}.
Although the issue of designing a desired collective behavior could be perceived as similar in many of these cases, the complexity of the design and assessment process varies as much as the diversity of the scenarios.
Depending on the research goals, differences arise in the underlying hypotheses guiding the development of the system, the opportunities to abstract intervening factors and isolate key study variables, and the ability to establish experimental protocols suitable for statistical analysis.

We illustrate the varying complexity of designing robot swarms using a conceptual framework borrowed from organizational theory and systems thinking~\cite{Ack1971MANSC}.
Specifically, we examine the issue from the perspective of three levels of complexity: \emph{puzzle}, \emph{problem}, and \emph{mess}.
Originally outlined by Ackoff in the 1970s~\cite{Ack1974book,Ack1981INTRF}, this conceptual framework helps identify and express the complexity of an issue based on the number of intervening factors, its general formulation, and available solutions.
In the following, we discuss how
(i)~the swarm robotics literature evolved by solving particular \emph{puzzles},
(ii)~recent advances in the automatic design of robot swarms are providing new tools to tackle a more general \emph{problem},
and (iii)~achieving large-scale robot swarms that operate in real-world scenarios is \emph{a mess}.
%
\section{THE PUZZLE}\label{sec:puzzle}

Ackoff defines \emph{puzzles} as issues in which the intervening factors and their relationships can be known and explicitly formulated.
Puzzles are well-structured and have solutions that can be identified through reasoning or principled approaches.
They are issues where all the necessary information can be made available and the solution becomes reproducible when the correct method is applied.
The puzzle exists within a closed system where the intervening factors can be isolated and studied independently; however, solving the puzzle may still require significant effort.

A \emph{puzzle} in the design of robot swarms is to find the set of rules and conditions that lead to the emergence of a particular collective behavior.
The aim is to understand underlying principles.
Typically, solving the puzzle involves manually applying a specific behavior model or a principled method to produce control software for the robots.
This research is described in a large part of the swarm robotics literature.
In the early years of swarm robotics, the puzzles helped to attract attention to the field with questions such as, how can a robot swarm aggregate?~\cite{GarJosGau-etal2008AL}, or how can division of labor emerge in a robot swarm?~\cite{KriBilKel2000NATU}.
In a puzzle, the swarm is seen as a closed system, and it is expected that a solution to the puzzle can be discovered with enough time, expertise and effort devoted to the design process.

A significant body of literature now demonstrates that self-organization is viable in autonomous groups of robots, supported by the application of behavior models and principled methods.
Recent examples include the emergence of shape~\cite{SunZhoMa-etal2023NATUCOM}, locomotion~\cite{LiBatBro-etal2019NATU}, and planning~\cite{GarBir2018SCIROB}.
Moreover, rather established taxonomies have characterized the diverse set of collective behaviors demonstrated~\cite{BraFerBirDor2013SI,SchUmlSenElm2020FRAI}.
Currently, addressing the design of robot swarms in the form of a puzzle helps unveil mechanisms of self-organization---for example, underwater coordination~\cite{BerGauNag2021SCIROB} or self-assembly under microgravity~\cite{NisCheMak-etal2022icra}.
%

\section{THE PROBLEM}\label{sec:problem}

In Ackoff's framework, a \emph{problem} is more complex than a puzzle.
Although a problem may still have identifiable solutions, they are not immediately apparent, and discovering them may require developing innovative approaches to assist the process.
Moreover, a problem often has multiple solutions, and finding the optimal one requires balancing preferences on various intervening factors.
A problem exists within a closed system, like a puzzle, but also requires handling uncertainties or incomplete information.

A more general \emph{problem} in the design of robot swarms is to systematically explore, select, customize, and combine sets of rules and conditions that enable self-organization, with the aim of controlling the emergence of diverse and tailored collective behaviors.
The complexity of the design process increases.
In a sense, solving the problem requires developing a single general approach to solving many diverse swarm robotics puzzles.
The design space to be explored is therefore much larger than that of a puzzle, and addressing this broader problem requires developing methods to aid in the generation of control software for robots, semi- or fully automatic~\cite{BirLigBoz-etal2019FRAI}.
To this end, literature focuses on developing automatic methods for designing robot swarms, rather than manually applying specific models or principled methods~\cite{BirLigHas2020NATUMINT}.
In this approach, the issue of designing robot swarms is turned into an optimization problem.
Given the specifications of a task for the swarm, an optimization process searches for suitable instances of control software that allow the robots to collectively perform the task~\cite{BirLigBoz-etal2019FRAI}.

In the design of robot swarms, the more general \emph{problem} arises as the field matures into an engineering discipline, focusing on questions such as how to develop automatic methods that generalize to various robot platfomrs and tasks?~\cite{KegGarHas-etal2024RAL,Tri2008book}, or how do automatic methods perform compared to manually producing control software for the robots?~\cite{FraBraBru-etal2015SI}.
Recent demonstrations include the use of neuroevolution~\cite{DuaCosGom-etal2016PLOSONE}, modular methods~\cite{SalGarBir2024COMMENG}, novelty search~\cite{HasLigBir2023SWEVO}, and surprise minimization algorithms~\cite{KaiHam2022IEEETR}.
%
%
These methods are meant to be task-agnostic, robust to performance variance, and capable of identifying a best solution among multiple potential ones.
Currently, addressing the design of robot swarms as a general problem leads to the development of methods that support the realization of robot swarms under predefined task requirements---for example, with the specification of tasks via demonstrations~\cite{GhaKucGarBir2023icra} and the on-board automatic generation of control software~\cite{JonWinHauStu2019AIS}.
Most of these approaches still consider the robot swarm as a closed system.
Thus, the design process mostly handles uncertainties caused by the swarm itself, within controlled limits.
%

\section{THE MESS}\label{sec:mess}

A \emph{mess} is the most complex and challenging issue.
Ackoff describes messes as systems of problems that are interconnected and interdependent.
The core issue cannot be clearly defined; neither can a straightforward strategy be outlined to address it satisfactorily.
In a mess, problems cannot be separated and solved individually; they must be understood and addressed as a whole.
Messes are characterized by ambiguity, complexity, and uncertainty, and their resolution requires a holistic, systems-thinking approach~\cite{Ack1981INTRF}.

We argue that designing large-scale robot swarms capable of operating out of the box in real-world settings is currently \emph{a mess}.
Swarm robotics is approaching real-world applications~\cite{DorTheTri2021PIEEE}.
A central focus remains in identifying suitable rules for self-organization, both in the form of a puzzle or a problem.
However, the design process now extends to other interconnected issues, functional and non-functional, which relate to how robot swarms will operate in the real world.

We illustrate the current complexity of designing robot swarms by highlighting some of the issues that have gained attention in recent years, how they interrelate, and the research questions they raise.

In the real world, a robot swarm will have to function as an open system, exposed to unpredictable interactions with its environment.
In this scenario, should a swarm operate as a static rule-following system, or should it be capable of learning and adapting?
A swarm operating in an open and unstructured world could be endowed with distributed situational awareness to rapidly and accurately understand its environment and act accordingly~\cite{JonMilSooHau2020AIS}.
This awareness could also enable the collective accumulation of information and experience, opening opportunities for the continuous optimization of the behavior of the swarm through lifelong social learning and cultural evolution~\cite{BreFon2021PTRSLSB,CamAlbFre-etal2021ASOC}.

In the real world, robots will not operate in isolation.
They will interact with simple machines, other robots (or swarms), animals, and humans to perform their tasks~\cite{RahCebObr-etal2019NATU}.
Therefore, when should a swarm engage and coordinate with other entities and when should it remain independent from them?
To be more effective, a swarm could tailor its interaction rules to engage with entities that populate the environment~\cite{AswPin2023iros,GarBir2024icra}---whether cooperatively or non-cooperatively.
However, safety and security measures may need to be implemented to protect both the swarm and others~\cite{HunHau2020NATUMINT,CasHarPenDor2021SCIROB}.

Robot swarms will transition from the laboratory to the real world only if the design process effectively embodies societal motivations and concerns.
Under these conditions, how can the design process address the diverse interests of relevant stakeholders?
Robot swarms will have to be developed under policies and regulations that oversee their societal and ecological impact~\cite{SwaIveHau2023tas,KinPorStr-etal2023}.
To foster societal trust and widespread adoption, advances in the design of robot swarms must be accompanied by the development of technologies to monitor, explain, and verify their actions~\cite{WilChaWin-etal2023tas,AlhAbdHau2022ants,NaiSooRam2024dars}.
This should be applicable both during normal operation and in the edge cases~\cite{KucLucAvr-etal2024icra}.
On the other hand, the design process should ensure that the swarms are cost-effective and allow for the implementation of financial strategies to generate economic value with the robots~\cite{SalLigBir2019PEERJCS,DorPacReiStr2024NATUREN}.

Addressing the aforementioned problems and questions individually is difficult, and integrating their solutions into a cohesive research and development framework will be even more challenging.
We contend that to solve \emph{the mess}, a significant effort must be devoted to rethinking research goals and hypotheses, and to developing new holistic approaches and frameworks to design robot swarms.
%

\section{CONCLUSION}\label{sec:conclussion}

Swarm robotics can contribute to transform current robotics solutions into large-scale services~\cite{YanBelDup-etal2018SCIROB}.
Research approaches, methods, and tools have evolved to support this ambitious goal~\cite{DorTheTri2021PIEEE}.
Currently, the issue of designing robot swarms can be approached with varying levels of complexity.
Using Akoff's conceptual framework, we illustrated how
(i)~solving the \emph{puzzles} helps understand the principles of self-organization,
(ii)~tackling the more general \emph{problem} enables the systematic design of robot swarms,
and (iii)~key open questions in designing real-world robot swarms can be found in the \emph{mess}.
We highlight relevant research to motivate further work in promising directions.
%






\bibliographystyle{IEEEtran}
\bibliography{demiurge-bib/definitions.bib,demiurge-bib/author.bib,demiurge-bib/address.bib,demiurge-bib/proceedings.bib,demiurge-bib/journal.bib,demiurge-bib/publisher.bib,demiurge-bib/series-short.bib,demiurge-bib/institution.bib,demiurge-bib/bibliography.bib,demiurge-bib/newbibliography.bib}

\end{document}